\documentclass[a4, 10 pt, conference]{ieeeconf} 
\IEEEoverridecommandlockouts                            
\usepackage{graphicx}
\usepackage{capt-of}
\usepackage{float}
\usepackage{subcaption}
\usepackage{siunitx}
\usepackage{colortbl}
\usepackage{xcolor}
\definecolor{lightgray}{rgb}{0.83, 0.83, 0.83}
\usepackage{multirow}
\usepackage{booktabs}
\usepackage[normalem]{ulem}

\usepackage{amsmath,amssymb,mathabx}
\usepackage{amsfonts,mathtools}
\usepackage{mathrsfs,bm,dsfont}
\DeclareMathOperator{\argmin}{argmin}

\newtheorem{remark}{Remark}

\usepackage[
placement=top,
angle=0,
color=red,
scale=1.1,
hshift=0,
vshift=-15
]{background}
\backgroundsetup{contents={- Preprint submitted to the IEEE Robotics and Automation Letters -}}

\title{\LARGE \bf
Tag-based Visual Odometry Estimation for Indoor UAVs  Localization
}

\author{Massimiliano Bertoni$^{1}$, Simone Montecchio$^{2}$, Giulia Michieletto$^{1}$, Roberto Oboe$^{1}$ and Angelo Cenedese$^{2}$
\thanks{*This work is partially supported 
by MUR through PRIN
Grant DOCEAT 2020RTWES4.}
\thanks{$^{1}$M.Bertoni, G.Michieletto, and R.Oboe are with the Department of Management and Engineering, University of Padova, Italy.
        }%
\thanks{$^{2}$ S.Montecchio and A.Cenedese are with the Department of Information Engineering, University of Padova, Italy.
        }%
\thanks{Contact:
{\tt massimiliano.bertoni@phd.unipd.it}
        }%
}

\begin{document}

\maketitle
\thispagestyle{empty}
\pagestyle{empty}

\begin{abstract}
The agility and versatility offered by UAV platforms still encounter obstacles for full exploitation in industrial applications due to their indoor usage limitations. A significant challenge in this sense is finding a reliable and cost-effective way to localize aerial vehicles in a GNSS-denied environment. In this paper, we focus on the visual-based positioning paradigm: high accuracy in UAVs position and orientation estimation is achieved by leveraging the potentials offered by a dense and size-heterogenous map of tags. In detail, we propose an efficient visual odometry procedure focusing on hierarchical tags selection, outliers removal, and multi-tag estimation fusion, to facilitate the visual-inertial reconciliation. Experimental results show the validity of the proposed localization architecture as compared to the state of the art.
\end{abstract}

\section{INTRODUCTION}

In the last decades, aerial robotics has emerged as a valuable technology, especially in outdoor scenarios as those referring to rural and civil applications~\cite{kim2019unmanned,shakhatreh2019unmanned}. Conversely, the employment of Unmanned Aerial Vehicles (UAVs) within the industry and manufacturing sector is still limited, although promising~\cite{elmeseiry2021detailed}. This is in part due to the need for high-performance navigation solutions that rest upon real-time, robust, and accurate localization strategies. This requirement becomes particularly stringent in indoor structured and possibly cluttered environments, where the positioning methods based on the Global Navigation Satellites System (GNSS) information often exhibit poor performance. This is primarily attributed to the insufficient reliability of the GNSS signal caused by physical obstructions and/or the sensor's limited accuracy relative to the dimensions of the working areas. Hence, alternative sources of position information are desirable for the UAVs indoor localization~\cite{gyagenda2022review}.

 Many well-stated indoor localization methods exploit onboard laser rangefinders~\cite{wen2022uav}, or external motion capture systems~\cite{leong2019vision}. However, these solutions generally entail high costs and elevated computational loads. Another branch of the existing literature is devoted to strategies based on Ultra Wideband (UWB) architectures, usually combined with Inertial Measurement Unit (IMU) or Inertial Navigation System (INS) devices (see e.g.~\cite{song2021tightly,you2020data}). The principal drawbacks of these approaches are their performance degradation caused by interference from conductive materials, and their limited working frequency (around $\SI{10}{Hz}$). Currently, the robotic community efforts are also focused on developing indoor localization solutions based on the machine learning paradigm~\cite{sandamini2023review}: however, even if promising and effective in many cases, these require a large amount of labeled training data and specialized, as well as energy-intensive, hardware.
 Finally, several state-of-the-art works 
 rest on the exploitation of the data recorded by onboard vision sensors, namely monocular/stereo/depth cameras~\cite{arafat2023vision}. 
Along this line, a popular and effective strategy for UAVs accurate localization relies on the detection and identification of some fiducial markers, such as tags.  This approach is principally exploited when coping with precise landing maneuvers (see e.g.~\cite{zhang2019enhanced} and the references therein). However, a Visual Inertial Odometry (VIO) localization method has recently been proposed in~\cite{bertoni2022indoor} guaranteeing the pose (position and attitude) estimation for any multi-rotor UAV during a complete flight, from the take-off to the landing phase. Inspired by~\cite{kayhani2019improved} and similar works, the solution described in~\cite{bertoni2022indoor} guarantees the navigation of aerial platforms in indoor environments through the fusion of the data derived from visual tags processing and IMU sensor measurements. The outlined strategy requires the presence of a map of tags suitably designed for limiting as much as possible the loss of updated pose information during the flight, especially in the case of trajectories with a variable altitude.

In this work, we develop a tag-based visual odometry estimation procedure that provides high UAV localization accuracy. This can be used as a standalone approach or within a VIO scheme as in~\cite{bertoni2022indoor}.
In doing this, the novel and original contributions can be summarized as
\begin{itemize}
    \item a \textit{tag hierarchical selection} method to better exploit the size-heterogeneity of the employed map;
    \item an \textit{outlier removal} procedure aiming at discerning reliable data from all collected UAV pose information;
    \item a \textit{multi-tags estimation fusion} process, which implies the exploitation of the data from multiple tags detected in the scene through the merging of all the correspondingly computed reliable UAV pose estimates.
\end{itemize}
As compared to the existing methods, these operations generally imply the reduction of both the mean value and standard deviation of the position estimation error, as confirmed by the experimental results. Moreover, with respect to~\cite{bertoni2022indoor} we also improve the orientation estimation, considerably reducing the estimation error in view of physical interaction tasks.

The rest of the paper is organized as follows. 
Section~\ref{sec:solution} focuses on the proposed tags processing procedure, whose effects on the UAVs localization are then discussed in Section~\ref{sec:validation}. 
Concluding remarks are drawn in Section~\ref{sec:conclusions}.


\begin{figure}[t]
\centering
\subfloat[\label{fig:map_modified}]{%
 \includegraphics[height=3.5cm]{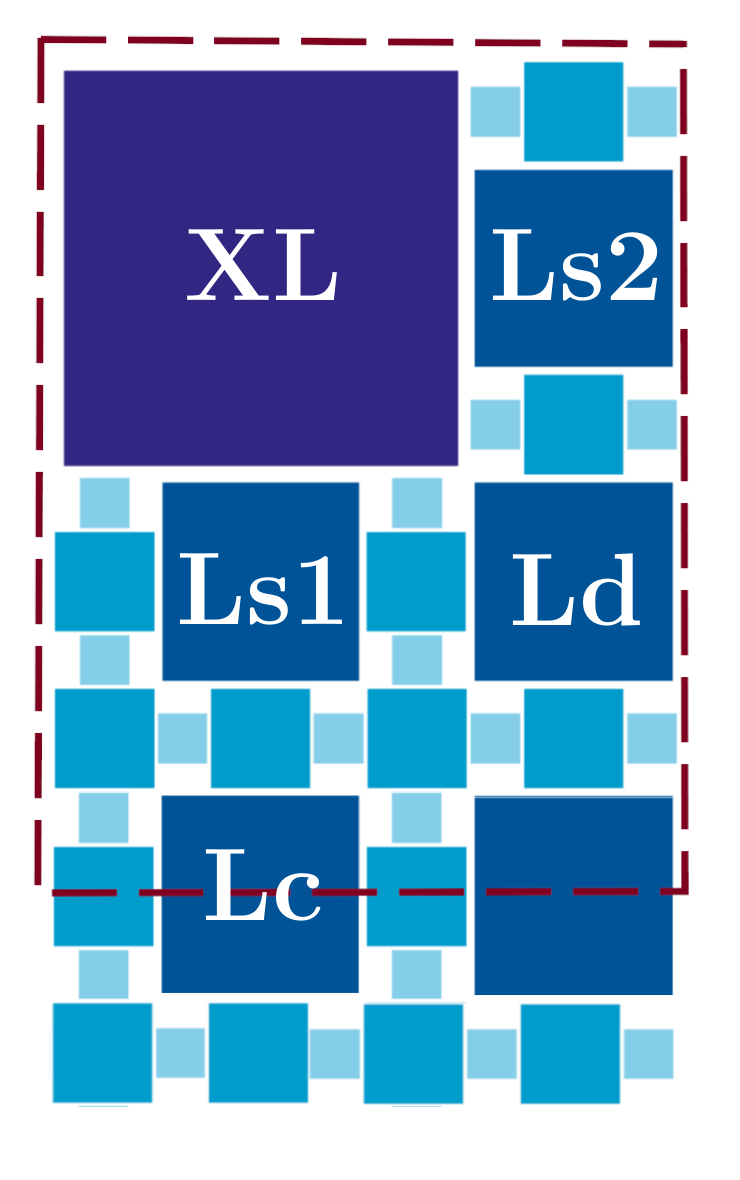 } 
}
\hfill
\subfloat[\label{fig:localization_overview_scheme}]{%
 \includegraphics[height=3.5cm]{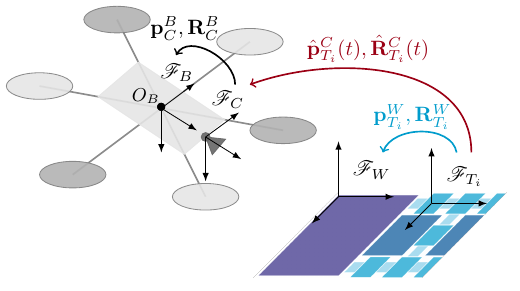}%
}
\caption{(a) portion of the tags map used for the lab experiments; (b) relationship among the involved reference frames.}
\label{fig:localization_overview}
\end{figure}

\section{TAG-BASED INDOOR LOCALIZATION}
\label{sec:solution}

In this technical section, the indoor UAVs localization problem is formally stated and the solution introduced in~\cite{bertoni2022indoor} is reviewed (Section~\ref{sec:problem}). Then, the original contributions of the work are provided, detailing the procedures to tackle the tags selection, the outliers removal, and the multi-tags estimation fusion (Section~\ref{sec:hierarchical},~\ref{sec:outliers} and~\ref{sec:datafusion}).


\subsection{Problem Statement and Background}
\label{sec:problem}

We focus on multi-rotor UAVs actuated by an arbitrary number $n \geq 4$ of propellers. 
Each of these UAVs is usually modeled as a rigid body acting in 3D space and, thus, it is characterized by a time-varying pose, meant as a position and an orientation. 
Formally, we consider a reference frame, named \textit{body frame} $\mathscr{F}_B$, in-build with the UAV so that its origin $O_B$ is centered on the vehicle CoM. Then, the vehicle pose coincides with the position of $O_B$ and the orientation of $\mathscr{F}_B$ with respect to a fixed reference frame $\mathscr{F}_W$ (\textit{world frame}). From a mathematical point of view, these are respectively identified by the vector $\mathbf{p}_B^W(t) \in \mathbb{R}^3$ and the rotation matrix $\mathbf{R}_B^W(t) \in SO(3)$. We assume that the matrix $\mathbf{R}_B^W(t)$ is a function of the time-varying roll $\phi(t) \in (-\pi,\pi]$, pitch $\theta \in (-\pi,\pi]$, and yaw angle $\psi(t) \in (-\pi,\pi]$ according to the rotation composition based on the ZYX sequence convention. We recall that   the matrix
 $\mathbf{R}_B^W(t)$ can also be interpreted as a function of the unit quaternion $\mathbf{q}_B^W(t) \in \mathbb{S}^3$.

Given these premises, we address the \textit{indoor UAVs localization problem}, presenting an effective solution to continuously compute a reliable estimation 
of the UAV pose during the vehicle flight, ensuring a high level of accuracy. In particular, we build on the strategy outlined in~\cite{bertoni2022indoor} with the purpose of reducing the error on the estimated position and orientation, i.e. on the estimated pose $\big(\hat{\mathbf{p}}_B^W(t),\hat{\mathbf{R}}_B^W(t)\big)\in \mathbb{R}^3 \times SO(3)$ with $\mathbb{R}^3 \times SO(3)= SE(3)$.

The VIO localization solution presented in~\cite{bertoni2022indoor} rests on the Kalman-based fusion of the information gathered by the onboard IMU sensors and a tag-based pose estimation. The latter involves the detection and identification of some fiducial markers composing a suitably designed map of tags. More specifically, the devised procedure collects images from an on-board camera, detects all the available tags in the scene, interprets the pose information in each marker, and computes the UAV pose at a high frame rate ($\sim$60 fps), exploiting an a-priori knowledge of the map. 
The map is specifically conceived so as to maximize the number of visible markers under various flying conditions and camera features in terms of field of view, resolution, and focus. 
Thus, it is characterized by a high number of narrowly placed tags belonging to different classes in terms of size and results from the roto-translation and mirroring operations of an optimized tags pattern. Figure~\ref{fig:map_modified} shows a schematic portion of the designed map: blue shades distinguish the tags sizes, and the red dashed line spotlights the optimized tags pattern.

Figure~\ref{fig:localization_overview_scheme} further clarifies the tag-based UAV localization. To ease the rigid body transformations, additional reference frames are introduced: the \textit{camera frame} $\mathscr{F}_C$ and the $i$-th \textit{tag frame} $\mathscr{F}_{T_i}$ with $i \in \{ 1 \ldots n_M\}$ being $n_M \in \mathbb{N}$ the cardinality of the markers composing the map. The camera frame is defined so that the $z$-axis is aligned with the camera focal axis, while the origin of $\mathscr{F}_{T_i}$ is assumed to be placed in the $i$-th tag geometrical center. Both the pose of the camera frame with respect to the body frame $\big(\mathbf{p}_C^B, \mathbf{R}_C^B\big) \in SE(3)$ and the pose of any $i$-th tag frame with respect to the world frame $\big(\mathbf{p}_{T_i}^W, \mathbf{R}_{T_i}^W\big) \in SE(3)$ are fixed over time and are supposed to be known.  
Exploiting such a piece of a-priori information, the vehicle is thus capable of estimating its pose with respect to the world frame, i.e., to compute $\big(\hat{\mathbf{p}}_B^W(t), \hat{\mathbf{R}}_B^W(t)\big)$, thanks to the estimation of the (time-varying) pose $\big(\hat{\mathbf{p}}_{T_i}^C(t), \hat{\mathbf{R}}_{T_i}^C(t)\big) \in SE(3)$ of any $i$-th detected marker in the camera frame. 

\begin{figure}[t]
\centering
 \includegraphics[width=0.98\columnwidth]{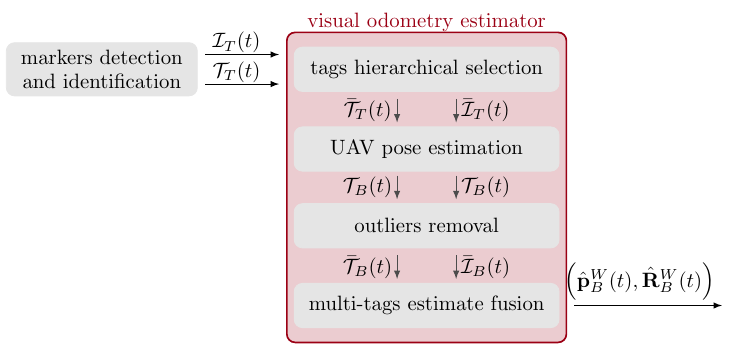}%
\caption{scheme of the newly proposed visual odometry procedure for the tag-based UAV pose estimation.}
\label{fig:localization_scheme}
\end{figure}


\subsection{Tags Hierarchical Selection}
\label{sec:hierarchical}

Figure~\ref{fig:localization_scheme} illustrates the scheme of the newly proposed visual odometry UAV pose estimation, which is here discussed highlighting the original contributions that allow improving the performance of the localization solution in~\cite{bertoni2022indoor}. 
The input of the depicted scheme consists of the set $\mathcal{I}_{T}(t)$ of the tags ID  and the set $\mathcal{T}_{T}(t)$ of tags estimated poses in the camera frame provided by a standard fiducial markers detection and identification method. The output, instead, corresponds to the UAV pose estimate $(\hat{\mathbf{p}}_B^W(t), \hat{\mathbf{R}}_B^W(t))$. 

The first step of the designed visual odometry procedure is a \textit{Tags Hierarchical Selection (THS)} which yields the definition of the sets $\widebar{\mathcal{I}}_{T}(t) \subseteq \mathcal{I}_{T}(t)$ and $\widebar{\mathcal{T}}_{T}(t) \subseteq \mathcal{T}_{T}(t)$, 
\begin{equation}
    \widebar{\mathcal{T}}_{T}(t) = \big\{ \big(\hat{\mathbf{p}}_{T_i}^C(t), \hat{\mathbf{R}}_{T_i}^C(t)\big) \in SE(3), \; i \in   \widebar{\mathcal{I}}_{T}(t)\big\}
\end{equation}
as detailed hereafter. 
We recall that in~\cite{bertoni2022indoor}, the computation of $\big(\hat{\mathbf{p}}_B^W(t), \hat{\mathbf{R}}_B^W(t)\big)$ is performed exploiting only the pose of the marker having the biggest size among those detected in the scene. 
In this work, instead, we aim at suitably exploiting the information from multiple tags to improve the UAV pose estimation accuracy. 
Thus, we distinguish the three following choices for the tag set selection:
\begin{itemize}
    \item the set $\widebar{\mathcal{I}}_{T}(t)$ is a singleton whose (single) element is the ID of the detected tag with the biggest size (hereafter, we refer to this selection as \textit{JBT - Just the Biggest Tag}); 
    \item the set $\widebar{\mathcal{I}}_{T}(t)$ is equal to ${\mathcal{I}}_{T}(t)$ if all the detected tags are taken into account (\textit{ALL} selection); 
    \item the set $\widebar{\mathcal{I}}_{T}(t)$ contains the ID of the detected tags belonging to the two markers classes with bigger size in the visible scene  (\textit{TBS - Two Biggest tags Sets} selection). 
\end{itemize}

The sets $\widebar{\mathcal{I}}_{T}(t)$ and $\widebar{\mathcal{T}}_{T}(t)$ then represent the input of the \textit{UAV Pose Estimation} step wherein the pose of the aerial platform is computed by exploiting the estimated pose of the selected tags. In detail, this step outputs the IDs set  ${\mathcal{I}}_{B}(t) = \widebar{\mathcal{I}}_{T}(t)$ and of the estimated poses collection
 \begin{equation}
     \mathcal{T}_{B}(t) = \big\{ \big(\hat{\mathbf{p}}_{i,B}^W(t), \hat{\mathbf{R}}_{i,B}^W(t)\big) \in SE(3), \; i \in   {\mathcal{I}}_{B}(t)\big\}.
 \end{equation}


\subsection{Outliers Removal}
\label{sec:outliers}

 Given the level of nuisance that typically affects mobile robots visual tasks, a subsequent \textit{Outliers Removal (OR)} step is introduced to identify the set $\widebar{\mathcal{I}}_{B}(t)  \subseteq \mathcal{I}_{B}(t)$ of tags ID  such that the corresponding set $ \widebar{\mathcal{T}}_{B}(t) \subseteq \mathcal{T}_{B}(t)$ contains the estimates of the UAV pose classified as reliable, thus being
 \begin{equation}
  \widebar{\mathcal{T}}_{B}(t) = \big\{ \big(\hat{\mathbf{p}}_{i,B}^W(t),\hat{\mathbf{R}}_{i,B}^W(t)\big) \!\in\! SE(3),  \; i \in \widebar{\mathcal{I}}_{B}(t) \big\}.
 \end{equation} 
 
To ease the notation, hereafter we drop the indication of reference frames when it is not necessary. Moreover, we focus only on the estimation of the UAV position, namely on the subset $\mathcal{P}_{B}(t)=\big\{ \hat{\mathbf{p}}_{i}(t) \in \mathbb{R}^3, i \in {\mathcal{I}}_{T}(t) \big\}  \subset {\mathcal{T}}_{B}(t)$, to efficiently identify the outliers in ${\mathcal{T}}_{B}(t)$. 

We rest on the well-known OR method based on Interquartile Ranges (IQR), which turns out to be suitable also for datasets whose distribution does not precisely fit a Gaussian one. 
The standard implementation of the method works for scalar quantities. Hence, we separately consider the components of the vectors in $\mathcal{P}_{B}(t)$, working on the sets
\begin{align}
\mathcal{P}_{k,B}(t) = \big\{&\hat{p}_{i,k}(t)\in \mathbb{R}  \, \vert \,  \\
&\hat{\mathbf{p}}_{i}(t) =\left[ \hat{p}_{i,1}(t) \; \hat{p}_{i,2}(t)\; \hat{p}_{i,3}(t)\right]^\top \in \mathcal{P}_{B}(t)\big\}  \nonumber
\end{align}
with $k \in \{1,2,3\}$.
For any $k$, assuming that $\mathcal{P}_{k,B}(t)$ contains at least three elements, 
the IQR range $\check{p}_k(t) \in \mathbb{R}$ is computed as the difference between the upper $\overline{p}_k(t) \in \mathbb{R}$ 
and lower quartile $\underline{p}_k(t) \in \mathbb{R}$, according to the standard IQR method. Then, the set of reliable estimates of the corresponding position component is defined as
\begin{align}
\label{eq:valid_pos}
  \widebar{\mathcal{P}}_{k,B}(t) &= \big\{ \hat{p}_{i,k} (t) \in \mathcal{P}_{k,B}(t) \; \vert \; \\
  & \qquad \underline{p}_k(t)-\delta\cdot \check{p}_k(t) < \hat{p}_{i,k} < \overline{p}_k(t)+\delta\cdot \check{p}_k(t)\big\}, \nonumber
  \end{align}
  where $\delta=1.5$ is the selected gain.
Therefore, we determine the ID of the tags involved in the definition of the set~\eqref{eq:valid_pos}: these constitute the set $\widebar{\mathcal{I}}_{k,B}(t)$. To conclude, the set $\widebar{\mathcal{T}}_{B}(t)$ of the reliable estimates of the UAV pose is computed starting from the corresponding  ID set obtained by intersection as
\begin{equation}
    \widebar{\mathcal{I}}_{B}(t) = \widebar{\mathcal{I}}_{1,B}(t) \cap \widebar{\mathcal{I}}_{2,B}(t) \cap \widebar{\mathcal{I}}_{3,B}(t).
\end{equation}

\subsection{Multi-tags Estimate Fusion}
\label{sec:datafusion}

A further tags processing act entails the fusion of reliable estimates of the UAV pose.
In detail, the purpose of the \textit{Multi-tags Estimate Fusion (MEF)} step is to compute $\big(\hat{\mathbf{p}}_{B}^W(t), \hat{\mathbf{R}}_{B}^W(t)\big)$ by suitably merging the elements of $\bar{\mathcal{T}}_{B}(t)$.
To do this, it is necessary to cope with both position and orientation data. Thus, we introduce the following sets having the same cardinality $\bar{n} \in \mathbb{N}$ of $\bar{\mathcal{I}}_{T}(t)$
\begin{align}
\bar{\mathcal{P}}_{B}(t)&=\big\{ \hat{\mathbf{p}}_{i,B}^W(t) \in \mathbb{R}^3, \; i \in \bar{\mathcal{I}}_{T}(t) \big\}  \subset \bar{\mathcal{T}}_{B}(t), \label{eq:position_data} \\
\bar{{\mathcal{R}}}_{B}(t)&=\big\{ \hat{\mathbf{R}}_{i,B}^W(t) \in SO(3), \; i \in \bar{\mathcal{I}}_{T}(t) \big\} \subset \bar{\mathcal{T}}_{B}(t) \label{eq:rotation_data}
\end{align}
and we proceed with a two-phase MEF step as follows.

\subsubsection{Position Data Fusion} to merge the position data in~\eqref{eq:position_data}, we resort to the standard Euclidean mean. Thus, the UAV position estimate $\hat{\mathbf{p}}_{B}^W(t)$ is computed as
\begin{equation}
\label{eq:pos_mean}
    \hat{\mathbf{p}}_{B}^W(t) = \frac{1}{\bar{n}} \sum_{i=1}^{\bar{n}} \hat{\mathbf{p}}_{i,B}^W(t).
\end{equation}

\subsubsection{Orientation Data Fusion}\label{sec:rotation_data_fusion} in view of the existing literature devoted to rotations mean, to merge the orientation data in~\eqref{eq:rotation_data}  we adopt the quaternion representation and exploit the \textit{Quaternion $L_2$-mean (QL2)} strategy, which represents a good trade-off between the accuracy on the resulting estimation and the algorithm computational complexity. 
In general, to compute the mean of a given set $\mathcal{Q} = \{\mathbf{q}_1 \ldots \mathbf{q}_{\bar{n}}\}$, such a method involves the solution of the optimization problem 
\begin{equation}
\label{eq:rotation_mean}
    {\mathbf{q}}^\star = \underset{\mathbf{q}\in\mathbb{S}^3}{\argmin} \sum_{i=1}^{\bar{n}} d(\mathbf{q}_i,\mathbf{q}),
\end{equation}
where $d(\mathbf{q}_i,\mathbf{q}) = \min\big\{\Vert \mathbf{q}_i-\mathbf{q}\Vert,\,\Vert\mathbf{q}_i+\mathbf{q}\Vert\big\} \in \mathbb{R}$ is the {quaternion $L_2$-metrics}.
Problem~\eqref{eq:rotation_mean} is addressed in~\cite{hartley2013rotation}, where a closed-form solution is presented given some conditions on the set $\mathcal{Q}$. Further elaboration is provided in the following.
For any quaternions pair $(\mathbf{q}_i,\mathbf{q}_j)$, $i,j \in \{1 \ldots \bar{n}\}, i \neq j$ in $\mathcal{Q}$, we indicate with $d_{\angle}\big(\mathbf{R}(\mathbf{q}_i),\mathbf{R}(\mathbf{q}_j)\big)$ the Riemannian distance between the rotation matrices associated to the quaternions in the pair. Specifically, it holds that $d_{\angle}\big(\mathbf{R}(\mathbf{q}_i),\mathbf{R}(\mathbf{q}_j)\big)= \big\Vert\log\big(\mathbf{R}(\mathbf{q}_i)\mathbf{R}(\mathbf{q}_j)^\top\big)\big\Vert \in \mathbb{R}$. 
When for any $i,j \in \{1 \ldots \bar{n}\}, i \neq j$ it is $d_{\angle}\big(\mathbf{R}(\mathbf{q}_i),\mathbf{R}(\mathbf{q}_j)\big) < \pi/2$  and the sign of any $\mathbf{q}_i$ is so that  $\Vert \mathbf{q} - \mathbf{q}_i \Vert < \Vert \mathbf{q} + \mathbf{q}_i \Vert $ given $\mathbf{q} \in \mathbb{S}^3$, the solution of~\eqref{eq:rotation_mean} can be computed as
\begin{equation}
\label{eq:minimum}
    {\mathbf{q}}^\star= \frac{\tilde{\mathbf{q}}}{\Vert\tilde{\mathbf{q}}\Vert}\quad\text{with}\quad \tilde{\mathbf{q}} = \sum_{i=1}^{\bar{n}}\mathbf{q}_i.
\end{equation}
The described conditions ensure that the minimum~\eqref{eq:minimum} is global and unique, avoiding the issues deriving from the double coverage property of quaternions (i.e., the fact that any quaternion $\mathbf{q} \in \mathbb{S}^3$ and its opposite represent the same rotation and thus, it holds that $\mathbf{R}(\mathbf{q})=\mathbf{R}(-\mathbf{q}) $).
To efficiently compute the UAV orientation estimate, we consider the outlined averaging strategy applied to the set $\bar{{\mathcal{Q}}}_{B}(t) =\{ \hat{\mathbf{q}}_{i,B}^W(t) \in \mathbb{S}^3 \; \vert \; \mathbf{R}(\hat{\mathbf{q}}_{i,B}^W(t))=\hat{\mathbf{R}}_{i,B}^W(t) \in \bar{{\mathcal{R}}}_{B}(t))\}$. 
Then, we convert the resulting quaternion $\hat{\mathbf{q}}_{B}^W(t)$, solution of~\eqref{eq:rotation_mean},  in the corresponding rotation matrix $\hat{\mathbf{R}}_{B}^W(t)$.

\begin{remark}\label{rmk:weights}
We perform the position and orientation data processing by introducing some weights in both the sums~\eqref{eq:pos_mean} and~\eqref{eq:minimum}. Specifically, the weights  $\{w_i \in \mathbb{R}, i \in \{1 \ldots \bar{n}\}\}$ are the same for the two pose components and are assigned according to the size of the $i$-th corresponding tag. Intuitively, the estimates deriving from markers having bigger sizes are weighted more since these tags are usually more clearly visible even at higher altitudes.
\end{remark}

\begin{remark}
After the MEF step, we apply a FIR filter on the last five pose estimates in order to mitigate error spikes while causing, at the same time, a negligible delay.
\end{remark}


\section{EXPERIMENTAL VALIDATION}
\label{sec:validation}

In this section, we first detail the setup and implementation of the conducted laboratory experimental campaign (Section~\ref{sec:exp_setup}). Then, we discuss the achieved results with the purpose of validating the various steps composing the proposed visual odometry estimation (Section~\ref{sec:static_tests}) and the whole tag-based localization procedure  (Section~\ref{sec:dynamic_tests}).

\begin{table*}
\centering
 \resizebox{0.98\textwidth}{!}{
\begin{tabular}{cl|c|cccc|cccc}
\toprule
\multicolumn{2}{c|}{\multirow{2}{*}{}} &  {\multirow{2}{*}{{JBT}}} & \multicolumn{4}{c}{THS and OR validation - choice by design: \bf{(W2+QL2)-MEF}} & \multicolumn{4}{|c}{MEF validation - choice by design:  \bf{TBS-OR}} \\
\cmidrule(lr){4-11}
& &                          &  ALL-notOR &  ALL-OR & TBS-notOR & \textbf{TBS-OR} & (W1+CL2)-MEF & (W2+CL2)-MEF  & (W1+QL2)-MEF & \textbf{(W2+QL2)-MEF}  \\  
\midrule
\multirow{5}{*}{$e_p$} & XL & 6.03 $\pm$ 4.11  & 18.03 $\pm$ 13.08  & 8.88 $\pm$ 6.45 &  4.71 $\pm$ 2.72 & \cellcolor{gray!20} \textbf{3.34 $\pm$ 1.76} & \textbf{3.20 $\pm$ 1.72} & 3.34 $\pm$ 1.76 & \textbf{3.20 $\pm$ 1.72} &  \cellcolor{gray!20} 3.34 $\pm$ 1.76 \\ 
& Ld & 4.74 $\pm$ 2.63 & 17.14 $\pm$ 12.21  & 9.11 $\pm$ 5.78 &  5.28 $\pm$ 2.69 & \cellcolor{gray!20} \textbf{2.95 $\pm$ 1.69} & \textbf{2.92 $\pm$ 1.48} & 2.95 $\pm$ 1.69 & \textbf{2.92 $\pm$ 1.48} & \cellcolor{gray!20} {2.95 $\pm$ 1.69} \\ 
 & Ls1 & 4.52 $\pm$ 2.21 &  17.80 $\pm$ 12.81  & 8.47 $\pm$ 5.97 & 4.59 $\pm$ 2.54 & \cellcolor{gray!20} \textbf{3.01 $\pm$ 2.00} & 3.45 $\pm$ 1.53 & \textbf{3.01 $\pm$ 2.00} & 3.45 $\pm$ 1.53 & \cellcolor{gray!20} \textbf{3.01 $\pm$ 2.00} \\ 
& Ls2 & 3.07 $\pm$ 1.51 & 18.50 $\pm$ 13.65  & 8.00 $\pm$ 5.61  & 3.26 $\pm$ 2.36 & \cellcolor{gray!20} \textbf{2.73 $\pm$ 1.76} & 3.61 $\pm$ 1.88 & \textbf{2.73 $\pm$ 1.76} & 3.60 $\pm$ 1.88 & \cellcolor{gray!20} \textbf{2.73 $\pm$ 1.76} \\  
& Lc & 8.57 $\pm$ 3.72 & 16.35 $\pm$ 8.09 & 7.07 $\pm$ 2.44 &  3.21 $\pm$ 1.55 & \cellcolor{gray!20} \textbf{3.05 $\pm$ 1.20} & 3.07 $\pm$ 1.26 & \textbf{3.03 $\pm$ 1.24} & 3.07 $\pm$ 1.25 & \cellcolor{gray!20} 3.05 $\pm$ 1.20 \\ 
\midrule
\multirow{5}{*}{$e_o$}& XL & 4.25 $\pm$ 1.20 & 7.82 $\pm$ 2.20 & 4.29 $\pm$ 1.05 &  \textbf{3.20 $\pm$ 0.40} & \cellcolor{gray!20} 3.24 $\pm$ 0.41 &  4.21 $\pm$ 6.73 & 3.90 $\pm$ 8.64 & 3.48 $\pm$ 0.44 & \cellcolor{gray!20} \textbf{3.24 $\pm$ 0.41} \\ 
& Ld & 3.58 $\pm$ 1.14 & 4.67 $\pm$ 2.20 & \textbf{3.45 $\pm$ 1.19} &  4.72 $\pm$ 0.98 & \cellcolor{gray!20} {4.00 $\pm$ 0.68} & 4.61 $\pm$ 6.70 & 4.50 $\pm$ 3.55 & \textbf{3.94 $\pm$ 0.72} & \cellcolor{gray!20} 4.00 $\pm$ 0.68 \\   
& Ls1 & 5.99 $\pm$ 0.63 & 4.95 $\pm$ 1.86 & \textbf{3.77 $\pm$ 1.03} & 5.06 $\pm$ 0.70 & \cellcolor{gray!20} {4.50 $\pm$ 0.63} & 6.04 $\pm$ 7.48 & 5.20 $\pm$ 7.36 & 5.17 $\pm$  0.63 & \cellcolor{gray!20} \textbf{4.50 $\pm$ 0.63} \\   
& Ls2 & \textbf{2.91 $\pm$ 0.50} & 7.43 $\pm$ 2.24 & 3.82 $\pm$ 1.02 &  {3.00 $\pm$ 0.37} & \cellcolor{gray!20} 3.19 $\pm$ 0.47 & 3.70 $\pm$ 5.45 & 3.91 $\pm$ 6.51 & \textbf{3.04 $\pm$ 0.54} & \cellcolor{gray!20} 3.19 $\pm$ 0.47 \\  
& Lc & 4.61 $\pm$ 0.80 & 4.84 $\pm$ 2.80 & 4.00 $\pm$ 0.58 & \textbf{3.84 $\pm$ 0.50} & \cellcolor{gray!20} 3.98 $\pm$ 0.45 & 4.07 $\pm$ 0.17 & 4.82 $\pm$ 1.58 & 4.05 $\pm$ 0.18 & \cellcolor{gray!20} \textbf{3.98 $\pm$ 0.44} \\ 
\bottomrule 
\end{tabular}
}
\caption{mnv$\pm$std of position [cm] and orientation [$^\circ$] estimation errors; best values are bold, proposed method is shaded.}
\label{tab:static_tests}
\end{table*}


\subsection{Experimental Setup and Implementation}
\label{sec:exp_setup}

The experimental campaign is carried out in the SPARCS laboratory (Dept. of Information Engineering, University of Padova) covering the floor of the $\SI{5}{m} \times \SI{7}{m} \times \SI{3}{m}$ flying arena with a $\SI{3}{m} \times \SI{5}{m}$ map of fiducial markers. This is composed of Apriltags belonging to the family TagStandard41h12 and divided into four different classes defined by the side length, namely S ($\SI{5.75}{cm}$), M ($\SI{11.5}{cm}$), L ($\SI{23}{cm}$), XL ($\SI{46}{cm}$), as shown in Figure~\ref{fig:map_modified}. Given that the side length doubles for any subsequent tag size, according to Remark~\ref{rmk:weights}, two sets of weights arise naturally: (W1) $w_i=4^h$ and (W2) $w_i=2^h$, with $h=0,1,2,3$ when the $i$-th tag belongs to the class having size S, M, L and XL, respectively. Note that the two proposed sets relate the sizes of the tags based on the proportion of the areas and the side length, respectively. 

We use the same multi-rotor UAVs as in~\cite{bertoni2022indoor}: a small-size quadrotor and a medium-size hexarotor having a diameter of $\SI{\sim0.3}{m}$ and $\SI{\sim0.8}{m}$, propellers included, respectively (hereafter, we refer to these vehicles as QR and HR, respectively). Both of them are equipped with a camera placed in front of the platform pointing to the ground and working at 30~fps, and with a Raspberry Pi 4B that acts as a companion computer.
This latter implements the ROS2 architecture that realizes the overall scheme in Figure~\ref{fig:localization_scheme},  providing the localization output at $\SI{\sim20}{Hz}$. Specifically, the \emph{markers detection and identification} block includes standard ROS2 nodes to control and provide the camera settings (\texttt{camera params}), acquire images in raw format (\texttt{raw images}), identify and detect the tags in the scene, providing the markers ID (\texttt{tags ID}) and their pose in $\mathscr{F}_C$ (\texttt{tags tf}).
The \emph{visual odometry estimation} block, instead, is realized through a single custom-made node.

The experimental setup includes also a ground working station for data logging and visualization and a VICON motion capture system (composed of 10 Vero 2.2 cameras), which provides the ground truth data for the UAVs localization, working at $\SI{100}{Hz}$ and having submillimeter accuracy.


\subsection{Visual Odometry Estimator Validation}
\label{sec:static_tests}

To first validate the tags processing procedure illustrated in Figure~\ref{fig:localization_scheme}, static hovering tests are conducted on the HR: we control the platform to fly at a given desired position with a constant orientation, and zero linear and angular velocities. 
We perform the tests in five different hovering positions on the map, while regulating the UAV at three different altitudes (specifically, $\SI{0.8}{m}$, $\SI{1.4}{m}$ and $\SI{2}{m}$ from the ground): 
this choice is motivated by the intent of investigating the performance of the THS, OR and MEF operations when changing the visible portion of the tags map.

Table~\ref{tab:static_tests}
reports the mean value (mnv) and the standard deviation (std) of the errors $e_p$ and $e_o$ on the UAV position and orientation estimates. These are respectively computed by averaging over time the quantities
\begin{equation}
   \big\Vert \hat{\mathbf{p}}^{\text{MC}}(t) - \hat{\mathbf{p}}^{\text{TL}}(t) \big\Vert \quad \text{and} \quad d_{\angle}\big(\hat{\mathbf{R}}^{\text{MC}}(t),\hat{\mathbf{R}}^{\text{TL}}(t)\big), 
\end{equation} 
denoting with $\big(\hat{\mathbf{p}}^{\text{MC}}(t), \hat{\mathbf{R}}^{\text{MC}}(t)\big) \in SE(3)$ the output of the VICON motion capture  system and with $\big(\hat{\mathbf{p}}^{\text{TL}(t)}
 \hat{\mathbf{R}}^{\text{TL}}(t)\big) \in SE(3)$ 
the result of the proposed tag-based visual odometry pose estimation. 
In the table, the different cases (hovering positions) are listed by row, and the first column reports the results corresponding to the JBT selection~\cite{bertoni2022indoor}.

Focusing on the THS and OR steps, on the left side of Table~\ref{tab:static_tests} we compare the estimation results when adopting the OR method of Section~\ref{sec:outliers} on the ALL and the TBS selections (respectively, ALL-OR and TBS-OR). 
Note that this second choice allows us to speed up the filtering procedure. 
Then, we consider the cases where the OR step is neglected in correspondence to the same ALL and TBS selections (ALL-notOR and TBS-notOR). 
To fairly compare the THS and OR strategies, the MEF operation is always performed accounting for the weights set W2 and the QL2 rotation averaging method~\eqref{eq:rotation_mean}. 
Remarkably, in ALL-notOR case the estimation accuracy is the lowest, also with respect to the JBT case, because of the presence of very noisy samples derived from the smaller tags having poor visibility. 
Indeed, we highlight that the absence of any OR procedure in correspondence to the ALL selection (ALL-notOR) implies a considerable position estimation error. 
Conversely, from the comparison between the ALL/TBS-notOR and the ALL/TBS-OR cases, the benefit of the OR procedure and the TBS selection appear to be well-motivated, since, in general, the TBS-OR method turns out to be the most effective approach to provide reliable UAV pose estimates even with respect to the JBT. 
Actually, it leads to a more precise final estimation for both position and orientation, although the improvement with respect to the other OR methods is greater for the error $e_p$ rather than $e_o$. This last fact fits with the typical priority devoted to the position rather than attitude estimation because of its influence on flight stability.

On the right side of the table, instead, the MEF method of Section~\ref{sec:datafusion} exploiting the quaternion $L_2$-mean method~\eqref{eq:rotation_mean} is validated against a different rotation averaging technique and with different choices of weights (see Remark~\ref{rmk:weights}). 
We consider the \textit{chordal $L_2$-mean} (CL2) method based on the chordal-length metric for rotation matrices~\cite{hartley2013rotation}, which typically underestimates the arc-length metric (related to QL2). Given a set $\mathcal{R} = \{ \mathbf{R}_1 \ldots \mathbf{R}_n\}$, the CL2 method involves the computation of the matrix $\mathbf{R} \in SO(3)$ that minimizes the cost function $\sum_{i=1}^{n} d_{c}\big(\mathbf{R}_i, \mathbf{R}\big)^2$, where $d_{c}\big(\mathbf{R}_i, \mathbf{R}\big)=\Vert \mathbf{R}_i - \mathbf{R} \Vert_F$ denotes the chordal distance resting on the Frobenious norm. A quaternion-based solution of the aforementioned minimization problem is provided in~\cite{markley2007averaging}. This relies upon the computation of the eigenvector associated with the maximum eigenvalue of the matrix $\mathbf{Q} = \sum_{i=1}^{n} w_i \,\mathbf{q}_i \mathbf{q}_i^\top $ where   $\mathbf{q}_i \in \mathbb{S}^3$ is the quaternion representation of $\mathbf{R}_i \in SO(3)$ and $w_i \in \mathbb{R}$ is the corresponding weight. 
To investigate the performance of the MEF techniques, we apply the TBS-OR procedure, leveraging the previously obtained results. 
We note that the exploitation of CL2 or QL2 rotation averaging strategy does not affect the accuracy of the UAV position estimation (up to numerical rounding approximation), while the mnv and the std of the error on the rotation estimate are slightly smaller when employing the QL2 method. {Also, the different weights set do not yield significant advantage in using either W1 or W2, hence the choice of the weights W2 refers to a more conservative choice, less sensitive to possible measurement noise.}
More interestingly, we highlight how the THS/OR choices, which are structural in the design, well affect the performance, while weighting/averaging selection appears more optional in this sense. In this latter ancillarity, we see an inherent robustness of the proposed solution.


\subsection{Visual-Based Localization Performance}
\label{sec:dynamic_tests}

We then assess the performance of the whole tag-based localization solution by considering the trajectories evaluated in~\cite{bertoni2022indoor}, namely a planar square path along which the UAV is required to keep a constant height from the ground (T1) and a sequence of vertical steps while maintaining the same position on the horizontal plane (T2).
In addition, we present and discuss the results with reference to a more complex 3D spline trajectory (T3) aiming at assessing the localization performance also in more dynamic scenarios.
Multiple experiments are performed with both the QR and the HR in correspondence to each desired path and the averaged pose errors are reported. These represent valuable performance indexes of the proposed localization strategy, which is implemented including the TBS-OR and the QL2-MEF steps, with the W2 weights choice (hereafter, we refer to this TBS-OR-(W2+QL2)-MEF simply as NEW).

\subsubsection{T1 - Planar Square Trajectory}
in Table~\ref{tab:T1}, we distinguish the four phases of the square maneuver of side $\SI{1.8}{m}$ during which the UAV keeps the constant altitude of $\SI{0.8}{m}$ from the ground. The F/B (Forward/Backward) phase corresponds to a movement along the negative/positive direction of the $y$-axis of the world frame, while the R/L (Right/Left) phase to a movement along the negative/positive direction of the $x$-axis. 
The major accuracy improvement is recorded in correspondence to the orientation estimation: both the mnv and the std of error $e_o$ are smaller as compared to the JBT case, especially for the HR. 
Also for the position estimation, the precision is mostly increased when the NEW localization strategy is adopted. The most notable exception is observed in correspondence to the L phase when executed by the HR: this performance worsening could be justified by a particularly unfavorable portion of the visible tags map.

\begin{table}[]
    \centering
     \resizebox{0.98\columnwidth}{!}{
    \begin{tabular}{cc|c|cccc}
        \toprule
        \multicolumn{3}{c}{}    & \multicolumn{4}{c}{Phase} \\
        \cmidrule(lr){4-7}
        \multicolumn{3}{c}{}      & F & R & B & L \\
       \midrule
         \multirow{4}{*}{$e_p$} &  \multirow{2}{*}{QR} &   JBT & ${7.40 \pm 2.63}$ & ${9.64 \pm 2.88} $ & ${6.38 \pm 2.99}$ & ${5.71 \pm 3.94}$ \\ 

        &   & NEW  & $\mathbf{7.23 \pm 1.86}$ & ${9.72 \pm \mathbf{1.40}} $ & ${8.00 \pm \mathbf{2.27}}$ & $\mathbf{2.62 \pm 1.13}$ \\ 
 \cmidrule(lr){2-7}
        
       & \multirow{2}{*}{HR} & JBT & ${6.78 \pm 3.38}$ & ${8.67 \pm 3.61} $ & ${5.53 \pm 2.21}$ & ${10.27 \pm 7.10}$ \\ 

         &   & NEW & ${6.96 \pm \mathbf{1.76}}$ & $\mathbf{8.45 \pm 2.96} $ & $\mathbf{5.00 \pm 1.98}$ & ${13.10 \pm 8.41}$ \\ 

       \midrule
         \multirow{4}{*}{$e_o$} & \multirow{2}{*}{QR} &   JBT & ${2.70 \pm 0.22}$ & ${2.29 \pm 0.34} $ & ${2.77 \pm 0.26}$ & ${3.09 \pm 0.24}$ \\ 

         &  & NEW  & $\mathbf{2.67 \pm 0.12}$ & ${2.41 \pm \mathbf{0.23}} $ & $\mathbf{2.66 \pm 0.21}$ & $\mathbf{2.92 \pm 0.15}$ \\
 \cmidrule(lr){2-7}
        
       & \multirow{2}{*}{HR} & JBT & ${6.43 \pm 2.32}$ & ${6.72 \pm 2.95} $ & ${5.42 \pm 1.87}$ & ${5.88 \pm 3.86}$ \\ 

         &   & NEW & $\mathbf{4.31 \pm 1.17}$ & $\mathbf{6.09 \pm 2.51} $ & $\mathbf{3.70 \pm 1.00}$ & $\mathbf{5.68 \pm 3.41}$ \\  

        \bottomrule
    \end{tabular}
    }
    \caption{T1 - mnv$\pm$std of position~[cm] and orientation~[$^\circ$] estimation errors; best values for NEW are bold.}
    \label{tab:T1}
\end{table}

\begin{table*}[t]
\begin{minipage}[b]{0.75\textwidth}
    \centering
        \resizebox{0.98\columnwidth}{!}{
    \begin{tabular}{cc|c|ccccccc}
    \toprule
    \multicolumn{3}{c}{}     & \multicolumn{7}{c}{Phase} \\
    \cmidrule(lr){4-10}
      \multicolumn{3}{c}{}         & S0 & A1 & A2 & A3   & D1 & D2 & D3 \\
    \midrule
    \multirow{4}{*}{$e_p$} & \multirow{2}{*}{QR} & JBT & ${4.67 \pm 1.66}$ & ${8.09 \pm 2.65} $ & ${10.17 \pm 4.96}$ & ${10.72 \pm 3.74}$ & ${9.29 \pm 3.98} $ & ${7.60 \pm 2.64} $ & ${5.34 \pm 1.38} $ \\ 
    & & NEW & $\mathbf{2.89 \pm 1.35}$ & $\mathbf{4.00 \pm 0.90} $ & $\mathbf{4.47 \pm 1.29}$ & $\mathbf{4.98 \pm 1.39}$ & $\mathbf{4.31 \pm 1.47} $ & $\mathbf{3.66 \pm 1.23} $ & $\mathbf{3.39 \pm 1.21} $ \\ 
\cmidrule(lr){2-10}
 & \multirow{2}{*}{HR} &   
   JBT & ${4.48 \pm 1.66}$ & ${6.60 \pm 1.66} $ & ${9.70 \pm 2.77}$ & ${12.85 \pm 4.96}$ & ${9.72 \pm 2.76} $ & ${7.23 \pm 1.72} $ & ${4.81 \pm 1.84} $ \\    
 &   & NEW & $\mathbf{1.65 \pm 0.68}$ & $\mathbf{2.88 \pm 1.06} $ & $\mathbf{3.87 \pm 1.18}$ & $\mathbf{5.41 \pm 1.92}$ & $\mathbf{3.93 \pm 1.60} $ & $\mathbf{2.85 \pm 0.83} $ & $\mathbf{1.88 \pm 0.75} $ \\ 
    \midrule
    \multirow{4}{*}{$e_o$} & \multirow{2}{*}{QR} & JBT & ${2.75 \pm 0.17}$ & ${2.54 \pm 0.36} $ & ${3.15 \pm 0.45}$ & ${2.59 \pm 0.28}$ & ${3.18 \pm 0.41} $ & ${2.44 \pm 0.43} $ & ${2.77 \pm 0.18} $ \\ 
    & & NEW & ${2.75 \pm \mathbf{0.15}}$ & ${2.64 \pm \mathbf{0.18}} $ & $\mathbf{2.92 \pm 0.23}$ & ${2.81 \pm \mathbf{0.23}}$ & $\mathbf{2.93 \pm 0.26} $ & ${2.60 \pm \mathbf{0.26}} $ & ${\mathbf{2.63} \pm 0.19} $ \\ 
  \cmidrule(lr){2-10}
& \multirow{2}{*}{HR} &   JBT & ${5.92 \pm 1.69}$ & ${6.23 \pm 0.96} $ & ${6.03 \pm 1.42}$ & ${6.24 \pm 1.70}$ & ${6.22 \pm 1.28} $ & ${6.43 \pm 0.90} $ & ${4.86 \pm 1.72} $ \\ 
&   & NEW & $\mathbf{3.90 \pm 0.64}$ & $\mathbf{4.36 \pm 0.45} $ & $\mathbf{3.45 \pm 0.68}$ & $\mathbf{3.75 \pm 0.73}$ & $\mathbf{3.63 \pm 0.66} $ & $\mathbf{4.45 \pm 0.41} $ & $\mathbf{3.57 \pm 0.66} $ \\ 
    \bottomrule
    \end{tabular}
    }
\end{minipage}\hfill
\begin{minipage}[b]{0.24\textwidth}
    \centering
    \resizebox{0.98\columnwidth}{!}{
    \begin{tabular}{cc|c|c}
    \toprule
\multicolumn{3}{c}{}&  3D spline \\
\multicolumn{3}{c}{} & trajectory \\
    \toprule
     \multirow{4}{*}{$e_p$} & \multirow{2}{*}{QR} & JBT & ${37.35 \pm 16.91}$ \\ 
& & NEW & $\mathbf{36.38 \pm 15.08}$ \\ 
    \cmidrule(lr){2-4}
 &     \multirow{2}{*}{HR} & JBT & ${12.65 \pm 7.78}$ \\ 
& &    NEW & $\mathbf{10.65 \pm 5.67}$ \\ 
    \midrule
         \multirow{4}{*}{$e_o$} & \multirow{2}{*}{QR} & JBT & ${3.01 \pm 3.04}$ \\ 
& & NEW & ${3.32 \pm 3.28}$ \\ 
    \cmidrule(lr){2-4}
 &     \multirow{2}{*}{HR} & JBT & ${5.74 \pm 2.91}$  \\ 
& &    NEW & $\mathbf{5.27 \pm 2.30}$ \\ 
\bottomrule
    \end{tabular}
    }
\end{minipage}
    \caption{T2[left] and T3[right] $\!\!$-$\!\!$ mnv$\pm$std of position~$\!$[cm] and orientation $\!$[$^\circ$] estimation errors; best NEW values are bold.}
    \label{tab:T2T3}
\end{table*}

\subsubsection{T2 - Vertical Steps Trajectory} Table~\ref{tab:T2T3} [left] reports on the tests conducted with the UAV flying over a tag having L size (Lc in Figure~\ref{fig:map_modified}) while accomplishing the takeoff maneuver (S0) and then an ascending (A$\bullet$) and a descending (D$\bullet$) sequence. Specifically, both the ascent and descent phases consist of three consecutive steps of amplitude $\SI{0.3}{m}$ so that the UAV altitude varies from $\SI{0.7}{m}$, when concluded the S0 phase, to $\SI{1.6}{m}$ from the ground (after A3 and before D1). 
We observe that by adopting the NEW localization solution the position estimation error is smaller in terms of mnv and std in correspondence to both the UAVs for all the phases. The same improvement is also recorded for the HR orientation estimation. For the QR, instead, we note that often only the mnv or only the std reduces.

\subsubsection{T3 - 3D Spline Trajectory} 
\begin{figure}[b!]
   \centering
   \includegraphics[width=1.00\columnwidth]{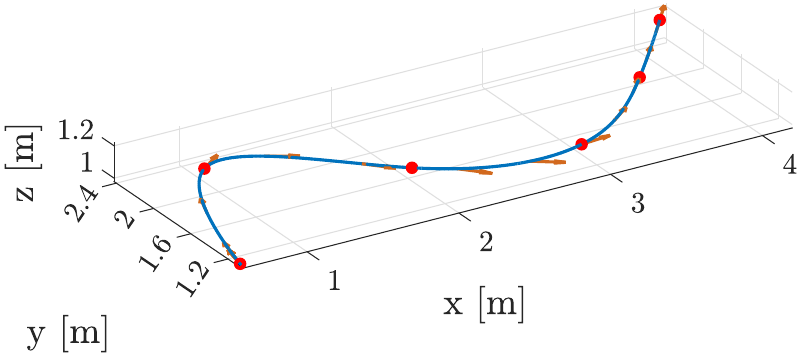}
   \caption{T3 - 3D spline trajectory with waypoints (red dots) and same samples of heading reference (orange arrows).}
   \label{fig:T3}
\end{figure}

the UAV is tasked to follow a path defined via polynomial interpolation of multiple waypoints. Figure~\ref{fig:T3} reports the resulting 3D trajectory along which the vehicle has also to track a varying yaw angle.
From Table~\ref{tab:T2T3}~[right], we note that the estimation error is smaller when adopting the JBT method only when accounting for the QR orientation. In all the other cases, both the mnv and the std are reduced when the NEW approach is in place.

\section{CONCLUSIONS}
\label{sec:conclusions}

Inspired by~\cite{bertoni2022indoor}, we propose a solution to indoor UAV localization based on a visual odometry estimation method that exploits an efficient tags processing procedure. This mainly acts in a threefold way. First, a hierarchical selection is performed on the markers detected in the scene. Then, the set of UAV pose estimates correspondingly calculated is purged based on an outlier removal IQR method. Finally, the reliable estimations are merged exploiting the weighted Euclidean mean for the position component and the quaternion-based rotation averaging technique for the orientation component. 

The efficacy of the proposed combination of the THS, OR, and MEF steps as compared to other possible ones is confirmed by the results of the conducted experimental campaign. The tests on the QR and HR platforms with both static positioning and dynamic trajectories have highlighted the estimation accuracy increase when adopting the proposed tag-based localization. Such performance enhancement is mainly evident in the case of static hovering, while both the maximum estimation error and the std are generally reduced for the dynamic trajectories.

In the future, to bridge the gap to real-world applications, we intend to investigate the localization performance in the presence of occlusions and/or a sparse map. In addition, we plan to deeply analyze the delays affecting the procedure and more thoroughly validate the performance on the more dynamic trajectories (e.g., T1, T3).


\bibliographystyle{IEEEtran}
\bibliography{references}

\begin{thebibliography}{10}
\providecommand{\url}[1]{#1}
\csname url@rmstyle\endcsname
\providecommand{\newblock}{\relax}
\providecommand{\bibinfo}[2]{#2}
\providecommand\BIBentrySTDinterwordspacing{\spaceskip=0pt\relax}
\providecommand\BIBentryALTinterwordstretchfactor{4}
\providecommand\BIBentryALTinterwordspacing{\spaceskip=\fontdimen2\font plus
\BIBentryALTinterwordstretchfactor\fontdimen3\font minus
  \fontdimen4\font\relax}
\providecommand\BIBforeignlanguage[2]{{%
\expandafter\ifx\csname l@#1\endcsname\relax
\typeout{** WARNING: IEEEtran.bst: No hyphenation pattern has been}%
\typeout{** loaded for the language `#1'. Using the pattern for}%
\typeout{** the default language instead.}%
\else
\language=\csname l@#1\endcsname
\fi
#2}}

\bibitem{kim2019unmanned}
J.~Kim, S.~Kim, C.~Ju, and H.~I. Son, ``Unmanned aerial vehicles in
  agriculture: A review of perspective of platform, control, and
  applications,'' \emph{IEEE Access}, vol.~7, pp. 105\,100--105\,115, 2019.

\bibitem{shakhatreh2019unmanned}
H.~Shakhatreh, A.~H. Sawalmeh, A.~Al-Fuqaha, Z.~Dou, E.~Almaita, I.~Khalil,
  N.~S. Othman, A.~Khreishah, and M.~Guizani, ``Unmanned aerial vehicles
  ({UAVs}): A survey on civil applications and key research challenges,''
  \emph{IEEE Access}, vol.~7, pp. 48\,572--48\,634, 2019.

\bibitem{elmeseiry2021detailed}
N.~Elmeseiry, N.~Alshaer, and T.~Ismail, ``A detailed survey and future
  directions of unmanned aerial vehicles ({UAVs}) with potential
  applications,'' \emph{Aerospace}, vol.~8, no.~12, p. 363, 2021.

\bibitem{gyagenda2022review}
N.~Gyagenda, J.~V. Hatilima, H.~Roth, and V.~Zhmud, ``A review of
  gnss-independent {UAV} navigation techniques,'' \emph{Robotics and Autonomous
  Systems}, p. 104069, 2022.

\bibitem{wen2022uav}
H.~Wen, W.~Nie, X.~Yang, and M.~Zhou, ``{UAV} indoor localization using {3D}
  laser radar,'' in \emph{IEEE 10th Asia-Pacific Conf. on Antennas and
  Propagation}.\hskip 1em plus 0.5em minus 0.4em\relax IEEE, 2022, pp. 1--2.

\bibitem{leong2019vision}
X.~W. Leong and H.~Hesse, ``Vision-based navigation for control of micro aerial
  vehicles,'' in \emph{4th IRC Conf. on Science, Engineering and
  Technology}.\hskip 1em plus 0.5em minus 0.4em\relax Springer, 2019, pp.
  413--427.

\bibitem{song2021tightly}
Y.~Song and L.-T. Hsu, ``Tightly coupled integrated navigation system via
  factor graph for {UAV} indoor localization,'' \emph{Aerospace Science and
  Technology}, vol. 108, p. 106370, 2021.

\bibitem{you2020data}
W.~You, F.~Li, L.~Liao, and M.~Huang, ``Data fusion of {UWB} and {IMU} based on
  unscented kalman filter for indoor localization of quadrotor {UAV},''
  \emph{IEEE Access}, vol.~8, pp. 64\,971--64\,981, 2020.

\bibitem{sandamini2023review}
C.~Sandamini, M.~W.~P. Maduranga, V.~Tilwari, J.~Yahaya, F.~Qamar, Q.~N.
  Nguyen, and S.~R.~A. Ibrahim, ``A review of indoor positioning systems for
  {UAV} localization with machine learning algorithms,'' \emph{Electronics},
  vol.~12, no.~7, p. 1533, 2023.

\bibitem{arafat2023vision}
M.~Y. Arafat, M.~M. Alam, and S.~Moh, ``Vision-based navigation techniques for
  unmanned aerial vehicles: Review and challenges,'' \emph{Drones}, vol.~7,
  no.~2, p.~89, 2023.

\bibitem{zhang2019enhanced}
X.~Zhang, J.~Jiang, Y.~Fang, X.~Zhang, and X.~Chen, ``Enhanced fiducial marker
  based precise landing for quadrotors,'' in \emph{IEEE/ASME Int. Conf. on
  Advanced Intelligent Mechatronics}.\hskip 1em plus 0.5em minus 0.4em\relax
  IEEE, 2019, pp. 1353--1358.

\bibitem{bertoni2022indoor}
M.~Bertoni, S.~Michieletto, R.~Oboe, and G.~Michieletto, ``Indoor visual-based
  localization system for multi-rotor {UAV}s,'' \emph{Sensors}, vol.~22,
  no.~15, p. 5798, 2022.

\bibitem{kayhani2019improved}
N.~Kayhani, A.~Heins, W.~Zhao, M.~Nahangi, B.~McCabe, and A.~P. Schoelligb,
  ``Improved tag-based indoor localization of {UAVs} using extended kalman
  filter,'' in \emph{Int. Symposium on Automation and Robotics in
  Construction}, 2019, pp. 21--24.

\bibitem{hartley2013rotation}
R.~Hartley, J.~Trumpf, Y.~Dai, and H.~Li, ``Rotation averaging,'' \emph{Int.
  Journal of Computer Vision}, vol. 103, pp. 267--305, 2013.

\bibitem{markley2007averaging}
F.~L. Markley, Y.~Cheng, J.~L. Crassidis, and Y.~Oshman, ``Averaging
  quaternions,'' \emph{Journal of Guidance, Control, and Dynamics}, vol.~30,
  no.~4, pp. 1193--1197, 2007.

\end{thebibliography}

\end{document}